\documentclass{article}
\usepackage{appendix}
\usepackage{spconf,amsmath,graphicx}
\usepackage{cite}
\usepackage{multirow}
\usepackage{multicol}
\usepackage{color}
\usepackage{amssymb,amsfonts,amsthm}
\usepackage{amsmath}               
  {
      \theoremstyle{plain}
      \newtheorem{assumption}{Assumption}
  }
\usepackage{algorithm}
\usepackage{algpseudocode}  
\usepackage{graphicx}
\usepackage[misc]{ifsym} 
\usepackage{subfigure} 
\usepackage{textcomp}
\usepackage{xcolor}
\usepackage{dsfont}
\usepackage{mathrsfs}
\usepackage{bbm}
\usepackage{booktabs} 
\usepackage{diagbox}
\usepackage{verbatim}
\usepackage{setspace}
\newtheorem{lemma}{Lemma}
\newtheorem{theorem}{Theorem}

\allowdisplaybreaks[4]
\title{Multi-Agent Adversarial Training Using Diffusion Learning}
\name{Ying Cao$^*$\qquad Elsa Rizk$^*$ \qquad Stefan Vlaski$^\dagger$ \qquad  Ali H. Sayed$^{*}$ 
\thanks{
\noindent Emails: ying.cao@epfl.ch, elsa.rizk@epfl.ch, s.vlaski@imperial.ac.uk, ali.sayed@epfl.ch}
}

\address{$^*$School of Engineering, École Polytechnique Fédérale de Lausanne\\
$^\dagger$ Department of Electrical and Electronic Engineering, Imperial College London\\}
\begin{document}
\ninept
\maketitle

\begin{abstract}
This work focuses on adversarial learning over graphs. We propose a general adversarial training framework for multi-agent systems using diffusion learning. We analyze the convergence properties of the proposed scheme for convex optimization problems, and illustrate its enhanced robustness to adversarial attacks.
\end{abstract}
\section{Introduction}\label{sec:intro}

In many machine learning algorithms, small malicious perturbations that are imperceptible to the human eye can cause classifiers to reach erroneous conclusions  \cite{szegedy2013intriguing,song2017pixeldefend,jia2017adversarial,pinto2017robust,sayed_2023}.  To mitigate the negative effect of adversarial examples, one methodology is adversarial training \cite{madry2017towards}, in which clean training samples are augmented by adversarial samples by adding purposefully crafted perturbations. Due to the lack of an explicit definition for the imperceptibility of perturbations, additive attacks  are usually restricted within a small bounded region. Most earlier studies, such as \cite{miyato2016adversarial,goodfellow2014explaining,madry2017towards,maini2020adversarial,zhang2019theoretically}, have focused on studying adversarial training in the context of single agent learning. In this work, we devise a robust training algorithm for multi-agent networked systems by relying on {diffusion learning \cite{sayed_2023,sayed2014adaptation,sayed2014adaptive}}, which has been shown to have a wider stability range and improved performance guarantees for adaptation in comparison to other decentralized strategies {\cite{sayed_2023,sayed2014adaptation,sayed2014adaptive}}.

There of course exist other works in the literature that applied adversarial learning to a multiplicity of agents, albeit using a different architecture. For example, the works \cite{qin2019adversarial,zhang2020distributed,liu2021concurrent} employ multiple GPUs and a fusion center, while the works \cite{feng2019graph,xu2019topology,wang2019graphdefense} consider graph neural networks. In this work, we focus on a fully decentralized architecture where each agent corresponds to a learning unit in its own right, and interactions occur locally over neighborhoods determined by a graph topology.

We formulate a sequential minimax optimization problem involving adversarial samples, and assume in this article that the perturbations are within an $\ell_2-$bounded region. We hasten to add though that the analysis can be extended to other norms, such as $\ell_1-$ and $\ell_{\infty}-$bounded perturbations. For simplicity, and due to space limitations, we consider the $\ell_2-$case here.

%adversarial training deals with the min-max optimization problem. In general, people study adversarial training on neural networks for which the inner maximization is not intractable. This can cause some unclear problems due to the inner approximation. For convex problems, we have closed-form solutions for the inner maximization, which makes it easier to study some fundamental problems in the field of robustness of machine learning models. This motivates us to study the multi-agent adversarial learning start from the convex case.

In the performance analysis, we examine the convergence of the proposed framework for convex settings due to space limitations, but note that similar bounds can be derived for nonconvex environments by showing convergence towards local minimizers. In particular, we show here that with strongly-convex loss functions, the proposed algorithm approaches the global minimizer 
within $O(\mu)$ after sufficient iterations, where $\mu$ is the step-size parameter.

\section{Problem Setting}
Consider a collection of $K$ agents where each agent $k$ observes independent realizations of some random data $(\boldsymbol{x}_k,\boldsymbol{y}_k)$,  where $\boldsymbol{x}_k$ plays the role of the {feature vector} and $\boldsymbol{y}_k$ plays the role of the {label variable}.  Adversarial training in the decentralized setting deals with the following stochastic minimax optimization problem
\begin{equation}\label{formu}
w^{\star} = \mathop{\mathrm{argmin}}\limits_{{w}\in \mathbbm{R}^M} \left\{J( w) \overset{\Delta}{=} \sum\limits_{k=1}^K \pi_k J_k( w)\right\}
\end{equation}\noindent where $\{\pi_k\}_{k=1}^{K}$ are positive scaling weights adding up to one, and each individual risk function is defined by
\begin{equation}\label{jk}
\begin{split}
    J_k(w) &= \mathds{E}_{\{\boldsymbol{x}_k,\boldsymbol{y}_k\}}\left\{ \max\limits_{\left\Vert{\delta}_k\right\Vert\le \epsilon}  Q_k(w;\boldsymbol{x}_k +{\delta}_k, \boldsymbol{y}_k)\right\} \\
\end{split}
\end{equation}
in terms of a loss function $Q_k(\cdot)$. In this formulation, the variable {${\delta}_k$ represents an $\ell_2$} norm-bounded perturbation used to generate adversarial examples, and $\boldsymbol{y}_k$ is the true label of sample $\boldsymbol{x}_k$. We refer to $w^{\star}$ as the robust model.  In this paper, we assume all agents observe data sampled independently (over time and space) from the same statistical distribution.

One methodology for solving (\ref{formu}) is to first determine the inner maximizer in (\ref{jk}), thus reducing the minimax problem to a standard stochastic minimization formulation. Then, the traditional stochastic gradient method could be used to seek the minimizer. We denote the true maximizer of the perturbed loss function in (\ref{jk}) by
\begin{align}\label{maximizer_t}
    \boldsymbol{\delta}_k^{\star} (w) \in \mathop{\mathrm{argmax}}\limits_{\left\Vert{\delta}_k\right\Vert  \le \epsilon}   Q_k(w;\boldsymbol{x}_k +{\delta}_k,\boldsymbol{y}_k)
\end{align}
where the dependence of $\boldsymbol{\delta}_k^\star$ on $w$ is shown explicitly. To apply the stochastic gradient method, we would need to evaluate the gradient of $Q(w;\boldsymbol{x}_k+\boldsymbol{\delta}_k^{\star}(w),\boldsymbol{y}_k)$ relative to $w$, which can be challenging since $\boldsymbol{\delta}_k^{\star}(w)$ is also dependent on $w$. {This difficulty} can be resolved by appealing to Danskin's theorem \cite{lin2020gradient, rockafellar2015convex, NEURIPS2019_05d0abb9}. Let 
\begin{align}
    g(w) \overset{\Delta}{=} \max_{\left\Vert{\delta}_k\right\Vert\le \epsilon} Q_k(w;\boldsymbol{x}_k+{\delta}_k,\boldsymbol{y}_k)
\end{align}Then, the theorem asserts that $g(w)$ is convex over $w$ if $Q_k(w;\cdot,\cdot)$ is convex over $w$. Moreover, $g(w)$ need not be differentiable over $w$ even when $Q_k(w;\cdot,\cdot)$ is differentiable. However, and importantly for our purposes, we can determine a subgradient for $g(w)$ by using the actual gradient of the loss evaluated at the worst perturbation, namely, it holds that
\begin{align}
\label{danskin}
    \nabla_w Q_k(w;\boldsymbol{x}_k+\boldsymbol{\delta}_k^\star,\boldsymbol{y}_k)\in \partial_w g(w)
\end{align}
where $\partial_{w}$ refers to the subdifferential set of its argument. In (\ref{danskin}), the gradient of $Q_k(\cdot)$ relative to $w$ at the maximizer $\boldsymbol{\delta}_k^{\star}$ is computed by treating $\boldsymbol{\delta}_k^{\star}$ as a stand-alone vector and ignoring its dependence on $w$. When $\boldsymbol{\delta}_k^{\star}$ in (\ref{maximizer_t}) happens to be unique, then the gradient on the left in (\ref{danskin}) will be equal to the right side, so that in that case the function $ g(w)$ is differentiable.

Motivated by these properties, and using (\ref{danskin}), we can now {propose} an algorithm to enhance the robustness of multi-agent systems to adversarial perturbations. To do so, we rely on the adapt-then-combine (ATC) version of the diffusion strategy  \cite{sayed2014adaptation,sayed2014adaptive} and write down the following adversarial extension to solve (\ref{formu})--(\ref{jk})
\begin{subequations}
\begin{align}
\label{newx_e}
&\boldsymbol{x}_{k,n}^{\star}= \boldsymbol{x}_{k,n}+\boldsymbol{\delta}^{\star}_{k,n}\\
\label{a2_e}
    &\boldsymbol{\phi}_{k,n} =\boldsymbol{w}_{k,n-1} - \mu\nabla_w Q_k(\boldsymbol{w}_{k,n-1};\boldsymbol{x}_{k,n}^{\star},\boldsymbol{y}_{k,n})\\
\label{a3_e}
&\boldsymbol{w}_{k,n}  = \sum\limits_{\ell \mathcal{\in N}_k} a_{\ell k} \boldsymbol{\phi}_{\ell,n}  
\end{align}
\end{subequations}
where 
\begin{equation}\label{max_k}
    \boldsymbol{\delta}^{\star}_{k,n} \in \mathop{\mathrm{argmax}}\limits_{\left\Vert{\delta}_k\right\Vert  \le \epsilon}   Q_k(\boldsymbol{w}_{k,n-1};\boldsymbol{x}_{k,n} +{\delta}_k,\boldsymbol{y}_{k,n})
\end{equation}In this implementation, expression (\ref{newx_e}) computes the worst-case adversarial example at iteration $n$ using the perturbation from $(\ref{max_k})$, while (\ref{a2_e}) is the intermediate adaptation step in which all agents simultaneously update their parameters with step-size $\mu$. Relation (\ref{a3_e}) is the convex combination step where the intermediate states $\boldsymbol{\phi}_{\ell,n}$ from the neighbors of agent $k$ are combined together. The scalars  $a_{\ell k}$ are non-negative and they add to one over $\ell \in {\cal N}_k$.

%Now that we know how to evaluate the gradient of the loss function using (\ref{danskin}), we still need to determine the true inner maximizer $\boldsymbol{\delta}^{\star}_{k,n}$ in (\ref{max_k}). This step will be discussed in the following for convex losses. 
\section{Convergence analysis}

This section analyzes the convergence of the adversarial diffusion strategy (\ref{newx_e})--(\ref{a3_e}) for the case of strongly convex loss functions. We list the following assumptions, which are commonly used in the literature of decentralized multi-agent learning and single-agent adversarial training \cite{sayed2014adaptation, sinha2017certifying, vlaski2019diffusion, vlaski2021distributed, vlaski2021distributed2}.

\begin{assumption}\label{as1}
    \textbf{(Strongly-connected graph)} The entries of the combination matrix $A = [a_{\ell k}]$ satisfy $a_{\ell k } \ge 0$ and the entries on each column add up to one, which means that $A$ is left-stochastic. Moreover, the graph is assumed to be strongly-connected, meaning that there exists a path with nonzero weights $\{a_{\ell k}\}$ linking any pair of agents and, in addition, at least one node $k$ in the network has a self-loop with $a_{kk}>0$.

\end{assumption}

\begin{assumption}\label{as3}

 (\textbf{Strong convexity}) For each agent k, the loss function $Q_k(w;\cdot)$ is $\nu-$strongly convex over $w$, namely, for any $w_1, w_2, x\in \mathbbm{R}^M$ and $y \in \mathbbm{R}$, it holds that
 \begin{align}\label{asconvex}
 Q_k(w_2;x,y) \ge \; & Q_k(w_1;x,y) + \nabla_{w^{\sf T}} Q_k(w_1;x,y)(w_2 - w_1) 
     \notag\\ 
     &+ \frac{\nu}{2}\Vert w_2 - w_1\Vert^2
 \end{align}
 
 $\hfill\square$
\end{assumption}

We remark that it also follows from Danskin's theorem \cite{lin2020gradient, rockafellar2015convex, NEURIPS2019_05d0abb9} that, when $Q_k(w;\cdot,\cdot)$ is $\nu-$strongly convex over $w$, then the adversarial risk {$J_k({w})$} defined by (\ref{jk}) will  be strongly convex. As a result, the aggregate risk $J(w)$ in (\ref{formu}) will be strongly-convex as well.

%Next, we require the loss function $Q_k$ to be smooth, which enables machine learning algorithms to be tractably computable \cite{sinha2017certifying}.

\begin{assumption}\label{as2}
 (\textbf{Smooth loss functions}): For each agent $k$, the gradients of the loss function relative to $w$ and $x$ are Lipschitz in relation to the three variables $\{w, x, y\}$ {in the following manner:}
\begin{footnotesize}
\begin{subequations}
\begin{align}
\label{smooth_q}
&\left\Vert\nabla_w  Q_k(w_2;x+\delta,y) -  \nabla_w  Q_k(w_1;x+\delta,y) \right\Vert\le   L_{1}\left\Vert w_2-w_1\right\Vert\\
\label{as2_4}
&\left\Vert\nabla_w  Q_k(w;x_2+\delta,y) - \nabla_w Q_k(w;x_1+\delta,y) \right\Vert\le L_{2}\left\Vert x_2-x_1\right\Vert\\
\label{as2_5}
&\left\Vert\nabla_w  Q_k(w;x+\delta,y_2) - \nabla_w Q_k(w; x+\delta, y_1) \right\Vert\le L_{3}\left\Vert y_2-y_1\right\Vert
\end{align}
\end{subequations}
\end{footnotesize}
and
\begin{footnotesize}
\begin{subequations}
\begin{align}
\label{as2_3}
&\left\Vert\nabla_x  Q_k(w_2;x+\delta,y) - \nabla_x  Q_k(w_1;x+\delta,y) \right\Vert\le L_{4}\left\Vert w_2-w_1\right\Vert\\
\label{as2_2}
&\left\Vert\nabla_x  Q_k(w;x_2+\delta,y) - \nabla_x  Q_k(w;x_1+\delta,y) \right\Vert\le L_{5}\left\Vert x_2-x_1\right\Vert
\end{align}
\end{subequations}
\end{footnotesize}where $\Vert\delta\Vert \le \epsilon$. For later use, we use $L = \max\{L_{1}, L_{2}, L_{3},L_4, L_5\}$.
\end{assumption}

\begin{assumption}\label{as4}
(\textbf{Bounded gradient disagreement}) For any pair of agents $k$ and $\ell$, the squared gradient disagreements are uniformly bounded on average, namely, for any $w\in \mathbbm{R}^M$ and $\Vert\delta\Vert\le \epsilon$, it holds that
\begin{equation}\label{as4_1}
    \mathds{E}_{\{\boldsymbol{x},\boldsymbol{y}\}} \Vert \nabla_w Q_k(w;\boldsymbol{x} + \delta,\boldsymbol{y}) - \nabla_w Q_{\ell}(w;\boldsymbol{x} + \delta,\boldsymbol{y})\Vert^2 \le C^2
\end{equation}
 $\hfill\square$
\end{assumption}

Note that (\ref{as4_1}) is automatically satisfied when all agents use the same loss function and collect data independently from the same distribution.  

To evaluate the performance of the proposed framework (\ref{newx_e})--(\ref{a3_e}), it is critical to compute the inner maximizer $\boldsymbol{\delta}_{k,n}^{\star}$ defined by (\ref{max_k}). Fortunately, for some convex problems, such as logistic regression, the maximization in (\ref{max_k}) has a unique closed-form solution, which will be shown in the simulation section. Thus, we analyze the convergence properties of (\ref{newx_e})--(\ref{a3_e}) when $\boldsymbol{\delta}_{k,n}^{\star}$ is unique. We first establish the following affine Lipschitz result for the risk function in (\ref{jk}); proofs are omitted due to space limitations.

\begin{lemma}\label{affine_l}
(\textbf{Affine Lipschitz})  For each agent $k$, the gradient of $J_k(w)$ is affine Lipschitz, namely, for any $w_1, w_2 \in \mathbbm{R}^M$, it holds that
\begin{equation}\label{affine_l_1}
    \Vert \nabla_w J_k(w_2) - \nabla_w J_k(w_1) \Vert^2 \le 2L^2\Vert w_2 - w_1 \Vert^2 + 8L^2\epsilon^2
\end{equation}
$\hfill\square$
\end{lemma}

Contrary to the traditional analysis of decentralized learning algorithms where the risk functions $J_k(w)$ are generally Lipschitz, it turns out {from (\ref{affine_l_1}) that} under adversarial perturbations, the risks in (\ref{jk}) are now {\em affine} Lipschitz. This requires adjustments to the convergence arguments. A similar situation arises, for example, when one studies the convergence of decentralized learning under non-smooth losses --- {see \cite{sayed_2023, ying2018performance1,ying2018performance2}}.

To proceed with the convergence analysis, we introduce the gradient noise process, which is defined by
\begin{equation}\label{d_gn}
  \boldsymbol{s}_{k,n}(\boldsymbol{w}_{k,n-1}) = \nabla_w Q_k(\boldsymbol{w}_{k,n-1};\boldsymbol{x}_{k,n}^\star,\boldsymbol{y}_{k,n}) - \nabla_w J_k(\boldsymbol{w}_{k,n-1}) 
\end{equation} This quantity measures the difference between the approximate gradient (represented by the gradient of the loss) and the true gradient (represented by the gradient of the risk). The following result establishes some useful properties for the gradient noise process, namely, it has zero mean and bounded second-order moment (conditioned on past history).
\begin{lemma}\label{p_gn}
(\textbf{Moments of gradient noise}) For each agent k, the gradient noise defined in (\ref{d_gn}) is zero mean and its variance satisfies
\begin{align}
   &\label{s0_convex} \mathds{E}\left\{\boldsymbol{s}_{k,n}(\boldsymbol{w}_{k,n-1})|\boldsymbol{\mathcal{F}}_{n-1}\right\}=0 \\
    &\label{s2_convex} \mathds{E}\left\{\left\Vert\boldsymbol{s}_{k,n}(\boldsymbol{w}_{k,n-1})\right\Vert^2|\boldsymbol{\mathcal{F}}_{n-1}\right\} \le 
            \beta_{k,\epsilon}^2 \left\Vert \widetilde{\boldsymbol{w}}_{k,n-1} \right\Vert^2 + \sigma_{k,\epsilon}^2
\end{align}for some non-negative scalars $\beta_{k,\epsilon}^2$ and $\sigma_{k,\epsilon}^2$ that depend on $\epsilon$ and can vary across agents. In the above notation, the quantity
 $\boldsymbol{\mathcal{F}}_{n-1}$ is the filtration generated by the past history of the random process $\mathrm{col}\{\boldsymbol{w}_{k,n}\}$, and
\begin{align}
   \widetilde{\boldsymbol{w}}_{k,n-1} = w^\star - {\boldsymbol{w}}_{k,n-1} 
\end{align}
$\hfill\square$
\end{lemma}
The main convergence result is stated next; the proof is again omitted due to space limitations.
\begin{theorem}\label{th_mse}
(\textbf{Network mean-square-error stability}) Consider a network of $K$ agents running the adversarial diffusion learning algorithm (\ref{newx_e})--(\ref{a3_e}). Under Assumptions \ref{as1}-- \ref{as4} and for sufficiently small step size $\mu$, {the network converges asymptotically to an $O(\mu)$-neighborhood around the global minimizer $w^{\star}$ at an exponential rate}, namely,
\begin{equation}
    \mathop{\lim \sup}\limits_{n\to\infty} \mathds{E}\left\Vert \widetilde{\boldsymbol{w}}_{k,n-1}\right\Vert^2 \le O(\mu)
\end{equation}
%with the convergence rate 
%\begin{align}
    %\lambda_{\epsilon} = 1 - 2\mu\nu + \mu^2 %b_{\epsilon}
%\end{align}
%Proof. Omitted due to space limitations.
$\hfill\square$
\end{theorem}
The above theorem indicates that the proposed algorithm enables the network to approach an $O(\mu)$-neighborhood of the robust minimizer $w^{\star}$ after enough iterations, so that the worst-case performance over all possible perturbations in the small region bounded by $\epsilon$ can be effectively minimized. %In other words, the robustness of the multi-agent systems can be improved.

\section{Computer Simulations}

In this section, we illustrate the performance of the proposed algorithm using a logistic regression application. Let $\boldsymbol{\gamma}$ be a binary variable that takes values from $\{-1,1\}$, and $\boldsymbol{h} \in \mathbbm{R}^M$ be a feature variable. The robust logistic regression problem by a network of agents employs the risk functions:
\begin{align}\label{18}
J_k (w) = \mathds{E}\max\limits_{\Vert{\delta}\Vert\le \epsilon}\left\{\ln{(1+e^{-\boldsymbol{\gamma}(\boldsymbol{h} + {\delta})^{\sf T}{w}}})\right\}
\end{align}
The analytical solution for the inner maximizer (i.e., the worst-case perturbation) is given by
\begin{align}
    \boldsymbol{\delta}^{\star} = -\epsilon\boldsymbol{\gamma}\frac{\boldsymbol{w}}{\Vert\boldsymbol{w}\Vert}
\end{align}
which is consistent with the perturbation computed from the fast gradient method (FGM) \cite{miyato2016adversarial}.
\begin{figure}[htbp]
\centerline{\includegraphics[width=6.5cm]{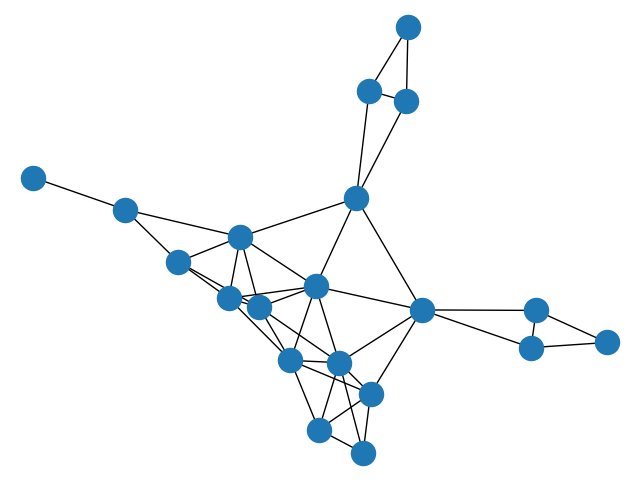}}
\caption{A randomly generated graph structure used in the simulations}
\label{fig1}
\end{figure}

\begin{figure}[htb]

\begin{minipage}[b]{1.0\linewidth}
  \centering
  \centerline{\includegraphics[width=7.5cm]{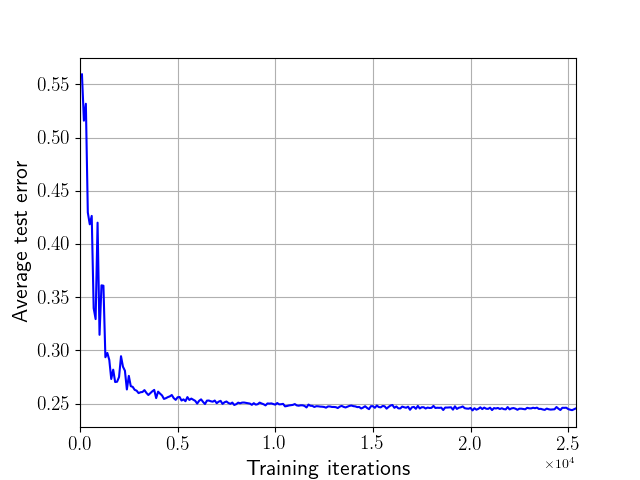}}
%  \vspace{2.0cm}
  \centerline{(a) MNIST}\medskip
\end{minipage}
\begin{minipage}[b]{1.0\linewidth}
  \centering
  \centerline{\includegraphics[width=7.5cm]{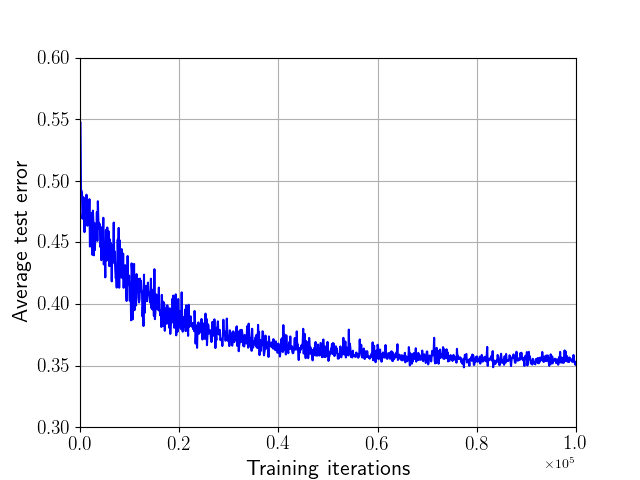}}
%  \vspace{1.5cm}
  \centerline{{(b) CIFAR10}}\medskip
\end{minipage}
%\hfill
\caption{The convergence plots for the two datasets: (a) The evolution of the average classification error over   adversarial examples bounded by $\epsilon = 4$ during training for MNIST; (b) The evolution of the average classification error of adversarial examples bounded by $\epsilon = 1.5$ during training for CIFAR10.}
\label{fig2}
\end{figure}

\begin{figure}[htb]

\begin{minipage}[b]{1.0\linewidth}
  \centering
  \centerline{\includegraphics[width=7.2cm]{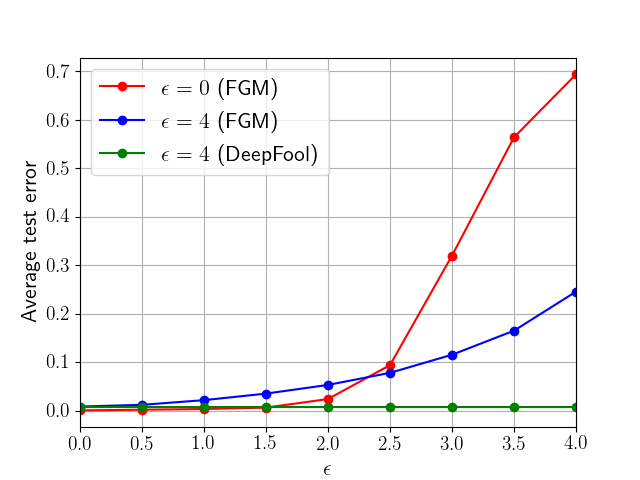}}
%  \vspace{2.0cm}
  \centerline{(a) MNIST}\medskip
\end{minipage}
\begin{minipage}[b]{1.0\linewidth}
  \centering
  \centerline{\includegraphics[width=7.2 cm]{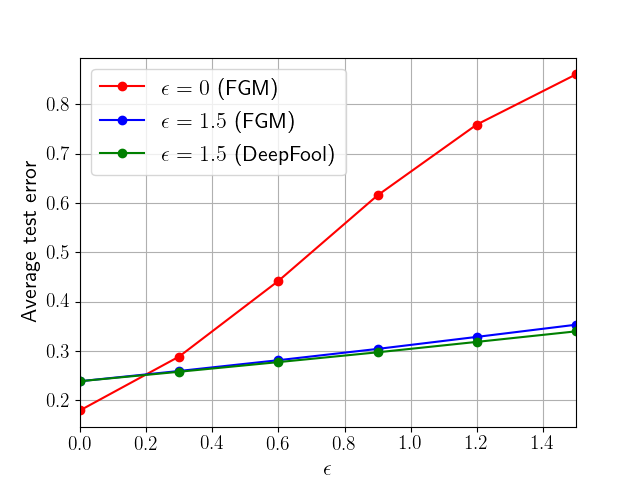}}
%  \vspace{1.5cm}
  \centerline{{(b)} CIFAR10}\medskip
\end{minipage}
\hfill
\caption{The robustness plots for the two datasets: (a) Classification error versus perturbation size for MNIST; (b) Classification error versus perturbation size for CIFAR10. The graphs show three plots illustrating the behavior of the traditional (nonrobust) algorithm to worst-case perturbations generated by means of FGM, as well as the performance of the proposed adversarial diffusion strategy (\ref{newx_e})--(\ref{a3_e}) to attacks generated by FGM and DeepFool.}
\label{fig3}
\end{figure}

\begin{figure}[htbp]
\centerline{\includegraphics[width=9cm]{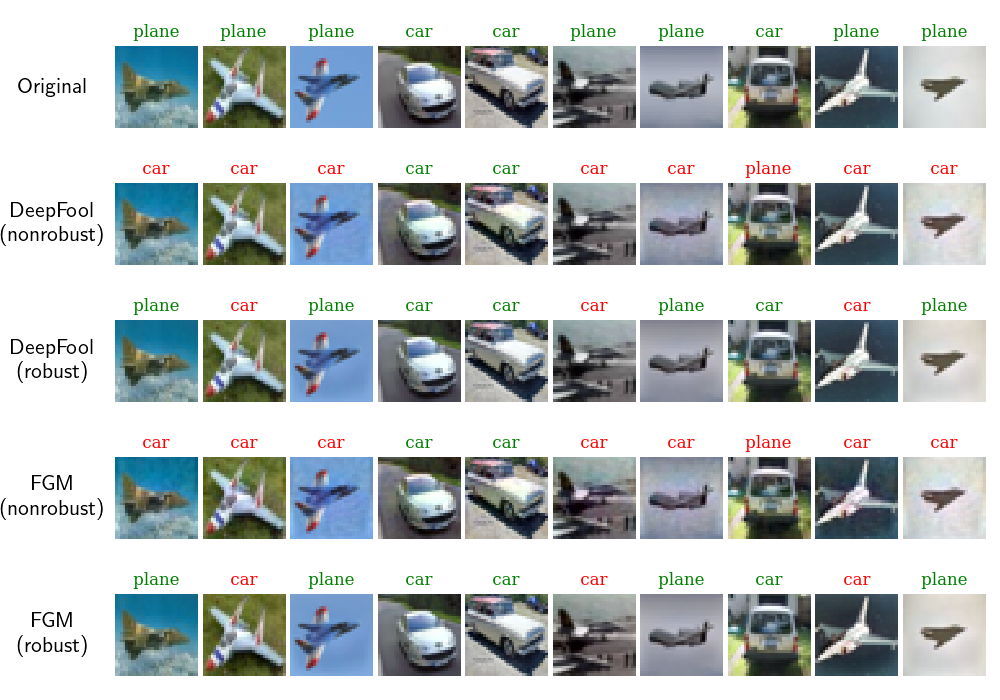}}
\caption{Visualization of the original and adversarial samples. The first row consists of 10 random original samples with the titles representing their true classes. The second row shows the adversarial examples generated by DeepFool and applied to the standard (nonrobust) algorithm. The third row shows the results obtained by the adversarial (robust) algorithm. The titles are the predictions by the corresponding models. The same construction is repeated in the last two rows using FGM. If the prediction of an image is wrong, the title is shown in red color. It is seen that the adversarial algorithm fails less frequently.} 
\label{fig4}
\end{figure}
In our experiments, we use both the MNIST \cite{deng2012mnist} and CIFAR10 \cite{krizhevsky2010cifar} datasets, and randomly generate a graph with 20 nodes, shown in Fig. \ref{fig1}. We limit our simulations to binary classification in this example. For this reason, we consider samples with digits \textit{$0$} and \textit{$1$} from MNIST, and images for airplanes and automobiles from CIFAR10. We set the perturbation bound in (\ref{18}) to $\epsilon = 4$ for MNIST and $\epsilon = 1.5$ for CIFAR10. In the test phase, we compute the average classification error across the network to measure the performance of the multi-agent system against perturbations of different strengths.

We first illustrate the convergence of our algorithm, as anticipated by Theorem \ref{th_mse}. From Fig. \ref{fig2}, we observe a steady decrease in the classification error towards a limiting value.

The robust behavior of the proposed algorithm is illustrated in Fig. \ref{fig3} for both MNIST and CIFAR10. We explain the curves for MNIST and similar remarks hold for CIFAR10. In the simulation, we use perturbations generated in one of two ways: using the FGM worst-case construction and also using the DeepFool construction {\cite{sayed_2023,moosavi2016deepfool}}. The figure shows three curves. The red curve is obtained by training the network using the traditional diffusion learning strategy without accounting for robustness. The network {is subsequently}  fed with worst-case perturbed samples (generated using FGM) during testing corresponding to different levels  of $\epsilon$. The red curve shows that the classification error deteriorates rapidly. The blue curve repeats the same experiment except that the network is now trained with the adversarial diffusion strategy (\ref{newx_e})--(\ref{a3_e}).  It is seen in the blue curve that the testing error is more resilient and the degradation is better controlled. The same experiment is repeated using the same adversarially trained network, where the perturbed samples are now generated using DeepFool as opposed to FGM. Here again it is seen that the network is resilient and the degradation in performance is better controlled.

In Fig. \ref{fig4}, we plot some randomly selected CIFAR10 images, their perturbed versions, and the classification decisions generated by the nonrobust algorithm and its adversarial version (\ref{newx_e})--(\ref{a3_e}). We observe from the figure that no matter which attack method is applied, the perturbations are always imperceptible to the human eye. Moreover, while the nonrobust algorithm fails to classify correctly in most cases, the adversarial algorithm is more robust and leads to fewer classification errors.

%\subsection{Neural networks}
%\begin{figure}[htbp]
%\centerline{\includegraphics[width=8.5cm]{errneural_multi_epsilon_fgm_3.png}}
%\caption{The evolution of the robustness of networks after training with different levels of attacks}
%\label{fig4}
%\end{figure}
%In this section, we show our algorithm improves the robustness of neural networks in distributed setting compared with the clean training. We use a randomly generated graph with 30 agents, where each agent is a neural network, and all agents work dependently according to the proposed algorithm. 
 
 %The experimental results are shown in Fig.\ref{fig4}, where we compare the robustness (i.e., classification error) of networks to different levels of $\ell_2$ attacks with $\epsilon$ ranging from 0 to 3 after training with $\epsilon = \{0,1,2,3\}$ respectively. Note that the curve corresponding to $\epsilon = 0$ means standard training, and the larger value of $\epsilon$ is equivalent to stronger attack. It is obvious that the clean network (i.e., trained with $\epsilon = 0$) Fig.\ref{fig4} is vulnerable to the crafted perturbations as its classification error goes up quickly when the attack becomes stronger. Fortunately, this phenomenon can be alleviated by our algorithm as
%the evolution of the classification error of the networks trained by $\epsilon  = \{1,2,3\}$ are more flat than the standard one. %What is more, in $\ell_2$ attacks with $\epsilon$ from $1$ to $3$, the robustness of networks is improved with the increase of $\epsilon$ used in the training process.

\section{Conclusion}

In this paper, we proposed a diffusion defense mechanism for adversarial attacks. We analyzed the convergence of the proposed method under convex losses and showed that it approaches a small $O(\mu)$ neighborhood around the robust solution. We further illustrated the behavior of the trained network to perturbations generated by FGM and DeepFool constructions and observed the enhanced robust behavior. Similar results are applicable to nonconvex losses and will be described in future work.

\bibliographystyle{IEEEbib.bst}
\bibliography{refs}

\end{document}